\documentclass{article}
\usepackage{ijcai19}

\usepackage{times}
\usepackage{soul}
\usepackage{url}
\usepackage[hidelinks]{hyperref}
\usepackage[utf8]{inputenc}
\usepackage[small]{caption}
\usepackage{graphicx}
\usepackage{amsmath}
\usepackage{amsfonts}
\usepackage{booktabs}
\usepackage[symbol]{footmisc}
\usepackage{algorithm}
\usepackage{algorithmic}
\usepackage{subfigure}
\usepackage{multirow}
\usepackage{color}
\newif\ifcomment
\commentfalse

\newcommand{\citet}[1]{\citeauthor{#1} \shortcite{#1}}

\title{Hybrid Actor-Critic Reinforcement Learning in Parameterized Action Space}

\author{
Zhou Fan\thanks{Equal contribution.}\and
Rui Su\footnotemark[1]\and
Weinan Zhang\thanks{Corresponding author.}\And
Yong Yu
\affiliations
Shanghai Jiao Tong University
\emails
zhou.fan@sjtu.edu.cn, \{surui, wnzhang, yyu\}@apex.sjtu.edu.cn
}

\begin{document}

\maketitle

\begin{abstract}
In this paper we propose a hybrid architecture of actor-critic algorithms for reinforcement learning in parameterized action space, which consists of multiple parallel sub-actor networks to decompose the structured action space into simpler action spaces along with a critic network to guide the training of all sub-actor networks. While this paper is mainly focused on parameterized action space, the proposed architecture, which we call hybrid actor-critic, can be extended for more general action spaces which has a hierarchical structure. We present an instance of the hybrid actor-critic architecture based on proximal policy optimization (PPO), which we refer to as hybrid proximal policy optimization (H-PPO). Our experiments test H-PPO on a collection of tasks with parameterized action space, where H-PPO demonstrates superior performance over previous methods of parameterized action reinforcement learning.
\end{abstract}

\section{Introduction}

Reinforcement learning (RL) has achieved impressive performance on a wide range of tasks including game playing, robotics and natural language processing. Most of recent exciting achievements is obtained by the combination of deep learning and reinforcement learning, known as deep reinforcement learning~\cite{mnih-atari-2013}. In game playing domains, deep Q-network (DQN)~\cite{mnih-atari-2013} is capable of learning control policies directly from high-dimensional sensory input in Atari games, and AlphaGo~\cite{alphago} has defeated world champions in the game of Go and could achieve superhuman performance even without human knowledge for training~\cite{alphago-2}. Robotics is also a significant aspect of applications of RL, where RL enables a robot to autonomously learn a sophisticated
 behavior through interactions with its environment ~\cite{rl-robotics-survey}. 

In the general setup of RL, an agent interacts with an environment in the following way: at each time step $t$, it observes  (either fully or partially) a state $s_t$ and takes an action $a_t$, then receives a reward signal $r_{t}$ as well as the next state $s_{t+1}$. Here the action $a_t$ is selected by the agent from its action space $\mathcal{A}$. The type of the action space is an important characteristic of the setup of an RL problem, and problems with different types of action space are usually solved with different algorithms. A typical RL setup may come with a discrete action space or a continuous one, and most RL algorithms are designed for either one of these two types. The agent simply selects its actions from a finite set of discrete actions if the action space is discrete, or from a single continuous space in the case of a continuous action space. However, action space could also have some hierarchical structure instead of being a flat set. The most common class of structured action space is known as parameterized action spaces, where a parameterized action is a discrete action parameterized by a continuous real-valued vector~\cite{q-pamdp}. With a parameterized action space, the agent not only selects an action from a discrete set, but selects the parameter to use with that action from the continuous parameter set of that action as well.

\begin{figure}[t]
\centering
\includegraphics[width=0.83\linewidth]{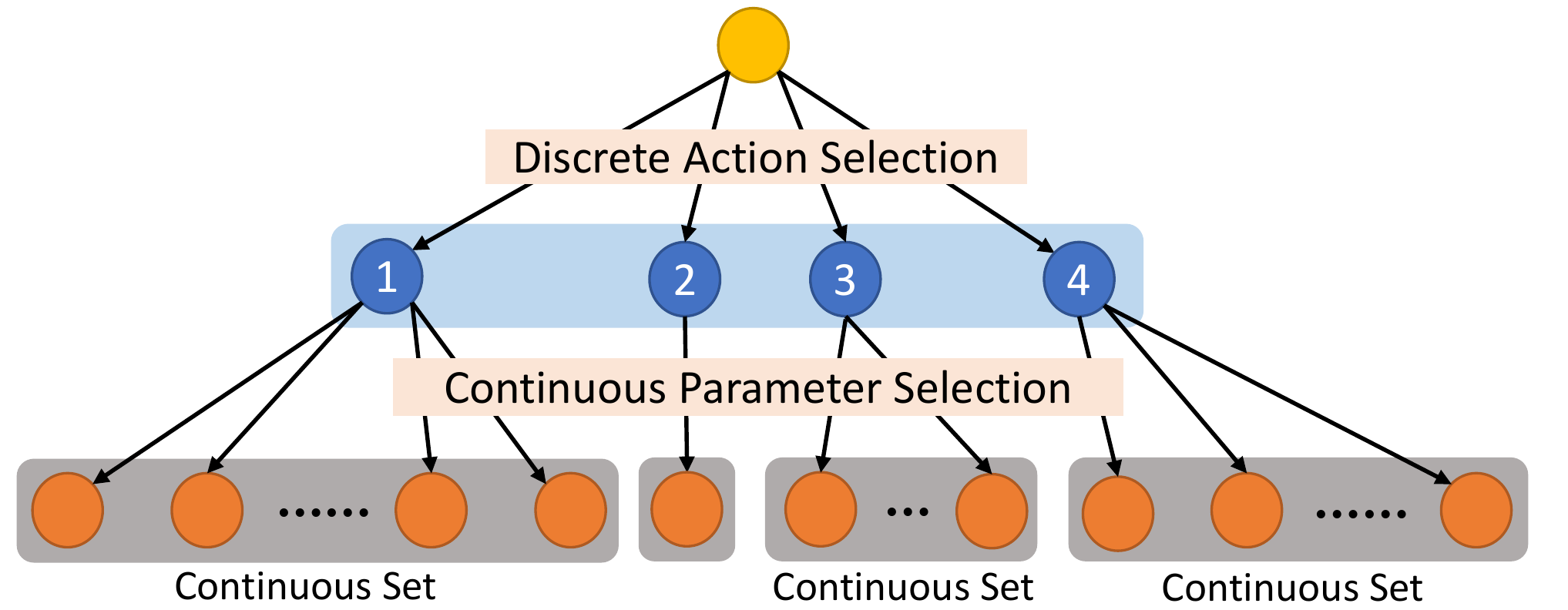}
\caption{Illustration of a parameterized action space.}
\label{fig:parameterized-action-space}
\end{figure}

Figure~\ref{fig:parameterized-action-space} shows an example of parameterized action space. The hierarchically structured action space contains four types of discrete actions shown in blue, and every discrete action has a continuous parameter space marked with rounded rectangles in grey. In this example, the discrete action with index $2$ is actually not parameterized, that is, there are no parameters to choose for this discrete action. It can also be viewed as a special case that the parameter space of discrete action $2$ only has one element. Parameterized action space perfectly models the scenarios where there are different categories of continuous actions. Many games as well as real world tasks have a parameterized action space. For example, in the Half Field Offense (HFO)~\cite{hfo} domain, which is a subtask based on the RoboCup 2D simulation platform, the agent may choose the discrete action Kick and specify its real-valued parameters (power and direction). Moreover, parameterized actions naturally exist in the context of robotics, where the action space can be constructed in a way that a set of meta-actions defines the higher-level selection of actions and every meta-action is controlled with fine-grained parameters \cite{rl-robotics-survey}. 

In addition to parameterized action spaces, action spaces may have more general hierarchical structures. For example, the parameters for the different actions are discretized in some game environments such as StarCraft II Learning Environment~\cite{star-craft-2}. Also, the action space may be manually constructed to have a hierarchical structure of more than two layers, which is a technique often used to reduce the size of an extremely large action space, with OpenAI Five on Dota 2 \cite{openai5} as a remarkable example. While it is intractable to choose an action directly from a set that contains millions of discrete actions, we can tackle this problem by constructing a hierarchically structured action space with a hierarchical taxonomy. 
\begin{figure}[t]
\centering
\includegraphics[width=0.85\linewidth]{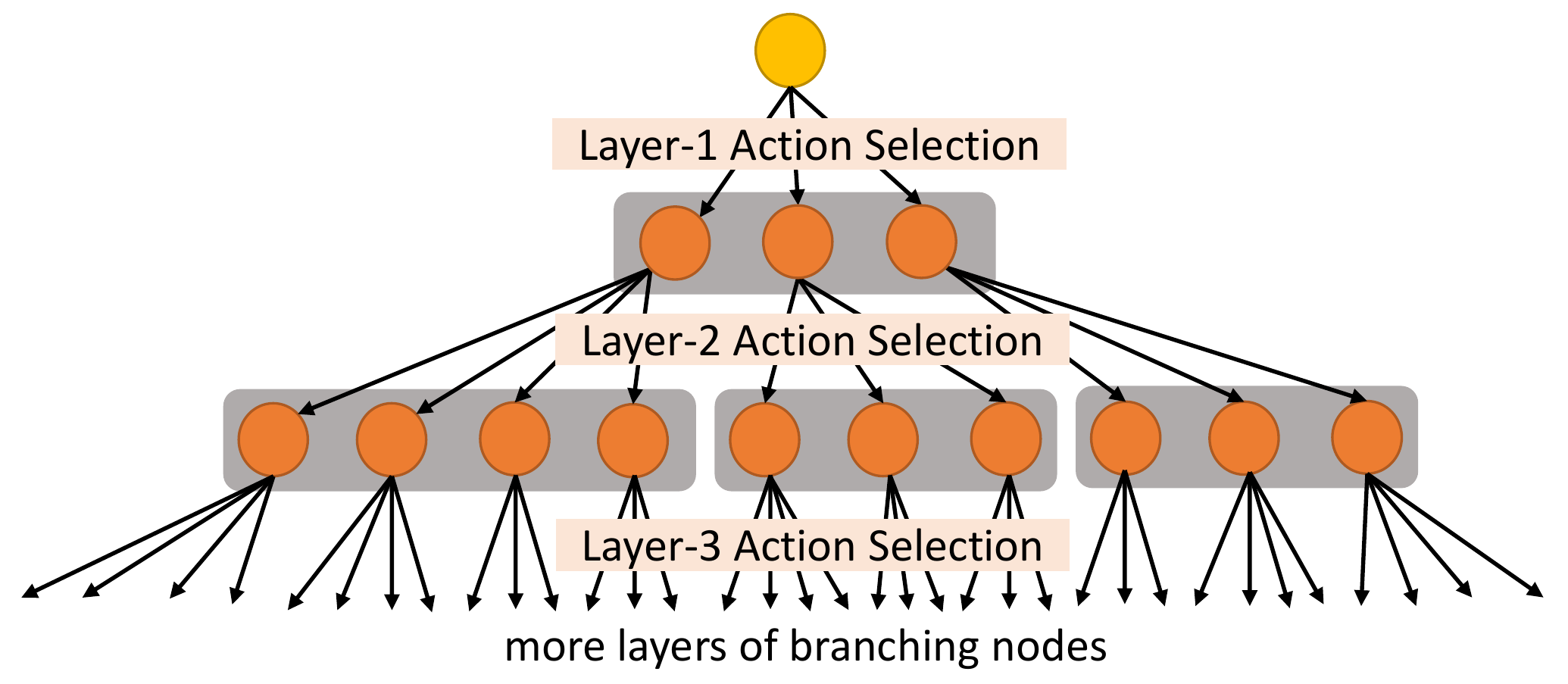}
\caption{A hierarchically structured action space.}
\label{fig:hierarchical-action-space}
\end{figure}
As is shown in Figure~\ref{fig:hierarchical-action-space}, the action space has a tree structure of multi-layer classifications of actions, in a way that each action selection node only has a small number of branches. Note that this tree structure could have more than two layers, and the external nodes of the tree structure could be continuous action-selection instead of discrete branching. In this view, the parameterized action space is a special case of hierarchical action space which has a discrete layer and then a continuous layer.

In this work, we propose a hybrid architecture of actor-critic algorithms for RL in parameterized action space. It is based on original architecture of actor-critic algorithms~\cite{actor-critic}, but contains multiple parallel sub-actor networks instead of one to solve multi-layer action selection respectively and has one global critic network to update the policy parameters of all sub-actor networks. Moreover, the hybrid actor-critic architecture we propose is flexible to the structure of the action space, such that it can also be generalized for other hierarchically structured action spaces. Specifically, we present an instance of the hybrid actor-critic architecture based on the proximal policy optimization (PPO)~\cite{ppo}, which we call hybrid proximal policy optimization (H-PPO). We show that H-PPO outperforms previous methods on a collection of tasks with parameterized action space.

The rest of this paper is organized as follows: Section~\ref{sec:related-work} introduces related work in the parameterized action space domain. The detailed architecture of hybrid actor-critic algorithms and H-PPO is presented in Section~\ref{sec:algorithm}. Section~\ref{sec:experiments} shows experiments and results. Finally, conclusion and future work is presented in Section~\ref{sec:conclusion}.

\section{Related Work} \label{sec:related-work}

Parameterized action spaces and other hierarchical action spaces are more difficult to deal with in RL compared to purely discrete or continuous action spaces for the following reasons. First, the action space has a hierarchical structure, which makes selecting an action more complicated than just choosing one element from a flat set of actions. Second, a parameterized action space involves both discrete action selection and continuous parameter selection, while most RL models are designed for only discrete action spaces or continuous action spaces. 

\subsection{RL Methods for Discrete Action Space and Continuous Action Space} 

The Q-learning algorithm~\cite{q-learning} is a value-based method which updates the Q-function using the Bellman equation
\begin{equation}
\label{equ:bellman-equation}
\resizebox{.91\linewidth}{!}{$
    \displaystyle
    Q(s, a) = \mathop{\mathbb{E}}_{r_t, s_{t+1}}[r_t + \gamma \max_{a' \in \mathcal{A}} Q(s_{t+1}, a') \mid s_t = s, a_t = a].
$}
\end{equation}
In the domain of discrete action space, deep Q-network (DQN)~\cite{mnih-atari-2013} takes the framework and uses a deep neural network to approximate the Q function. Some variations of DQN are also widely used in discrete action space, including asynchronous DQN~\cite{async-drl}, double DQN~\cite{double-dqn} and dueling DQN~\cite{dueling-q}. 

Policy gradient~\cite{policy-gradient} is another class of RL algorithms which optimizes a stochastic policy $\pi_{\theta}$ parameterized by $\theta$ to maximize the expected policy value $J(\pi_{\theta})$. The gradient of the stochastic policy is given by the policy gradient theorem~\cite{policy-gradient} as
\begin{equation}
\label{equ:policy-gradient-theorem}
    \nabla_{\theta} J(\pi_{\theta}) = \mathop{\mathbb{E}}_{s, a}[\nabla_{\theta}\log \pi_{\theta}(a \mid s)Q^{\pi_{\theta}}(s,a)].
\end{equation}
As an alternative, the policy gradient could also be computed with the advantage function $A^{\pi_{\theta}}(s, a)$ as
\begin{equation}
\label{equ:policy-gradient-advantage}
    \nabla_{\theta} J(\pi_{\theta}) = \mathop{\mathbb{E}}_{s, a}[\nabla_{\theta}\log \pi_{\theta}(a \mid s)A^{\pi_{\theta}}(s, a)].
\end{equation}

Similarly in continuous action spaces, the deterministic policy gradient (DPG) algorithm~\cite{dpg} and the DDPG algorithm~\cite{ddpg} optimize a deterministic policy $\mu_{\theta}$ parameterized by $\theta$ based on the deterministic policy gradient theorem~\cite{dpg} as
\begin{equation}
\label{equ:dpg-theorem}
    \nabla_{\theta} J(\mu_{\theta}) = \mathop{\mathbb{E}}_{s}[\nabla_{\theta}\mu_{\theta}(s)\nabla_{\theta}Q^{\mu_{\theta}}(s,a)\mid_{a=\mu_{\theta}(s)}].
\end{equation}

Based on the policy gradient methods, trust region policy optimization (TRPO)~\cite{trpo} and proximal policy optimization (PPO)~\cite{ppo} improve the optimization techniques to achieve better performance.

\subsection{RL Methods for Parameterized Action Space} 

To deal with the fact that a parameterized action space contains both discrete actions and continuous parameters, one straightforward approach is to directly discretize the continuous part of the action space and turn it into a large discrete set (for example with the tile coding approach~\cite{tile-coding}). This trivial method loses the advantages of continuous action space for fine-grained control, and often ends up with an extremely large discrete action space. 

Another direction is to convert the discrete action selection into a continuous space. \citet{parameterized-ddpg} used an actor network to output a value for each of the discrete actions, concatenated with all continuous parameters, and the discrete action is chosen to be the one with the maximum output value. The actor network is learned using the DDPG algorithm. By relaxing the structured action space into a continuous set, this method might significantly increase the complexity of the action space~\cite{pdqn}.

\citet{q-pamdp} focused on how to learn an action-selection policy given fixed parameter-selection, and proposed the framework called Q-PAMDP, which alternately learns the discrete action selection with Q-learning and updates parameter-selection policies with policy search methods.

\citet{hierarchical-approach-for-parl} proposed a hierarchical approach for RL in parameterized action space where the parameter policy is conditioned on the discrete action policy and used TRPO and Stochastic Value Gradient~\cite{svg} to train such an architecture. Although they also found that this method could be unstable due to the joint-learning between the discrete action policy and parameter policy.

\citet{pdqn} proposed the parameterized deep Q-networks (P-DQN) algorithm, which can be viewed as a combination of DQN and DDPG. P-DQN has one network to select the continuous parameters for all discrete action. Another network takes the state and the chosen continuous parameters as input  and outputs the Q-values for all discrete actions. The discrete action with the largest Q-value is chosen. However, the network that selects continuous parameters are updated to maximize the sum of the Q-values for all discrete actions, which might cause the algorithm being updated to improve the sum of the Q-values but decrease the largest Q-value. 

\section{Methodologies} \label{sec:algorithm}

This section introduces the proposed hybrid actor-critic architecture and presents the H-PPO algorithm as an instance of this architecture. Following the notations in \cite{q-pamdp}, we describe the parameterized action space in a mathematical way. We consider the following parameterized action space: the discrete actions are selected from a finite set $\mathcal{A}_d=\{a_1, a_2, \dots, a_k\}$, and each $a \in \mathcal{A}_d$ has a set of real-valued continuous parameters $\mathcal{X}_a \subseteq \mathbb{R}^{m_{a}}$. In this way, a complete action is represented as a tuple $(a, x)$, where $a \in \mathcal{A}_d$ is the chosen discrete action and $x \in \mathcal{X}_a$ is the chosen parameter to execute with action $a$. The whole action space $\mathcal{A}$ is then the union of each discrete action with all possible parameters for that action:
\begin{equation}
\label{equ:p-action-space}
    \mathcal{A} = \bigcup_{a \in \mathcal{A}_d} \{(a, x) \mid x \in \mathcal{X}_a\}.
\end{equation}
A Markov decision process with a parameterized action space is referred to as parameterized-action Markov decision processes (PAMDPs) \cite{q-pamdp}.

To design an algorithm for PAMDPs, we should first tackle the problem that parameterized action spaces are a class of discrete-continuous hybrid action spaces. Our method is based on the actor-critic architecture for the reason that many algorithms with actor-critic style, including policy gradient methods and PPO, are capable of learning stochastic policies in both discrete action spaces and continuous action spaces.
\begin{figure}[t]
\centering
\includegraphics[width=0.8\linewidth]{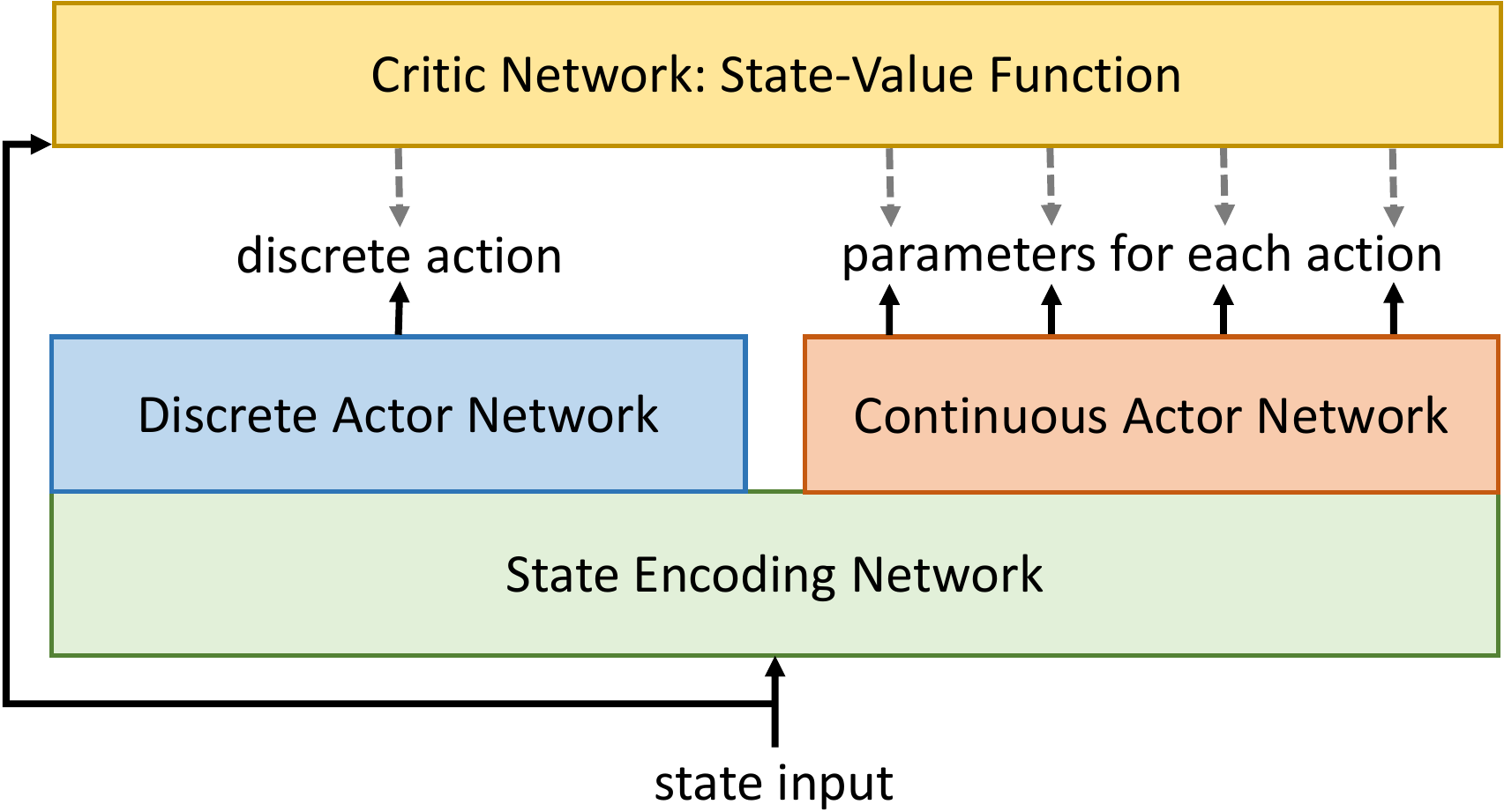}
\caption{Hybrid actor-critic architecture for parameterized action space.}
\label{fig:algorithm-structure-parameterized}
\end{figure}
Actor-critic algorithms usually have one actor network and one critic network, and the critic network is used to compute the gradient of the parameters of the actor network. By contrast, our proposed architecture for parameterized action space (shown in Figure~\ref{fig:algorithm-structure-parameterized}) contains two parallel actor networks (or even more for general hierarchical action spaces, introduced later in subsection~\ref{subsec:general-hybrid-actor-critic}). The parallel actors perform action-selection and parameter-selection separately: one discrete actor network learns a stochastic policy $\pi_{\theta_d}$ to select the discrete action $a$ and one continuous actor network learns a stochastic policy $\pi_{\theta_c}$ to choose the continuous parameters $x_{a_1}, x_{a_2}, \dots, x_{a_k}$ for all discrete actions. The complete action to execute is the selected action $a$ paired with the chosen parameter $x_{a}$ corresponding to action $a$. The two actor networks shares the first few layers to encode the state information. We refer to the proposed architecture as hybrid actor-critic architecture since discrete actor and continuous actor both exist in this architecture.

There is a single critic network in the hybrid actor-critic architecture, which works as an estimator of the state-value function $V(s)$. One important reason for us to use the state-value function as the critic instead of the action-value function is that action-value function suffers from the over-parameterization problem in parameterized action space. Specifically, if the action-value function is used as the critic, the critic network in implementation would take the state $s$, the selected discrete action $a$ and the chosen parameters for all discrete actions $x_{a_1}, x_{a_2}, \dots, x_{a_k}$ as input. It is impossible to just feed the chosen parameter $x_a$ for one specific discrete action $a$ into the critic network since the parameter dimensions of different discrete actions could be different. In this way, the action-value function is represented in the form of $Q(s, a, x_{a_1}, x_{a_2}, \dots, x_{a_k})$. However, the actual action to execute is not influenced by irrelevant parameters, so the true Q-function value is independent of $x_{a'}$ for all $a' \ne a$. Therefore, the action-value function would suffer from the problem of over-parameterization that 
\begin{equation}
\label{equ:q-value-equal}
    Q(s, a, x_{a_1}, x_{a_2}, \dots, x_{a_k}) = Q(s, a, x_{a}).
\end{equation}
By contrast, the state-value function only takes the state $s$ as input and does not have this problem. In our architecture, the state-value function $V(s)$ is used for computing a variance-reduced advantage function estimator $\hat{A}$. We follow the implementation used by \citet{async-drl}, which runs the policy for $T$ timesteps and computes the estimator $\hat{A}_t$ using the collected samples as
\begin{equation}
\label{equ:advantage-estimator}
\resizebox{.91\linewidth}{!}{$
    \displaystyle
    \hat{A}_t = -V(s_t) + r_t + \gamma r_{t+1} + \dots + \gamma^{T-t-1}r_{T-1} + \gamma^{T-t}V(s_T),
$}
\end{equation}
where $t \in [0, T]$ is the timestep index and $T$ is much less than the length of an episode. 

With a critic network providing estimation of the advantage function, the hybrid actor-critic architecture is flexible in the choice of the policy optimization method. The only requirement is that the optimization method should have an actor-critic style and updates stochastic policies with the advantage function provided by the critic. Although the complete action to execute $(a, x_a)$ is decided by both of the actors, the discrete actor and the continuous actor are updated separately by their respective update rules at each timestep. The update rules of the two actor networks could follow policy gradient methods as Eq.~(\ref{equ:policy-gradient-advantage}) or other optimization methods for stochastic policies such as PPO. We can even use two different optimization methods for the discrete policy $\pi_{\theta_d}$ and the continuous policy $\pi_{\theta_c}$. 

Then we present the hybrid proximal policy optimization algorithm for parameterized action space, which is a specific instance of the hybrid actor-critic architecture based on PPO.

\subsection{Hybrid Proximal Policy Optimization} \label{subsec:hppo}

The hybrid proximal policy optimization (H-PPO) takes the hybrid actor-critic architecture in Figure~\ref{fig:algorithm-structure-parameterized} and uses PPO as the policy optimization method for both its discrete policy $\pi_{\theta_d}$ and its continuous policy $\pi_{\theta_c}$. 

PPO is a state-of-the-art policy optimization method that learns a stochastic policy $\pi_{\theta}$ by minimizing a clipped surrogate objective~\cite{ppo} as
\begin{equation}
\label{equ:ppo-clip-objective}
\resizebox{.91\linewidth}{!}{$
    \displaystyle
    L^{\mathrm{CLIP}}(\theta) = \hat{\mathbb{E}}_t [\min(r_t(\theta)\hat{A}_t, \mathrm{clip}(r_t(\theta), 1-\epsilon,1+\epsilon)\hat{A}_t)],
$}
\end{equation}
where $r_t(\theta)$ denotes the probability ratio $r_t(\theta) = \frac{\pi_{\theta}(a_t \mid s_t)}{\pi_{\theta_{\mathrm{old}}}(a_t \mid s_t)}$ and $\epsilon$ is a hyperparameter.

To generate the stochastic policy for discrete actions $\pi_{\theta_d}$, the discrete actor network of H-PPO outputs $k$ values $f_{a_1}, f_{a_2}, \dots, f_{a_k}$ for the $k$ discrete actions, and the discrete action $a$ to take is randomly sampled from the $\mathrm{softmax}(f)$ distribution. The continuous actor network of H-PPO generates the stochastic policy $\pi_{\theta_c}$ for continuous parameters by outputting the mean and variance of a Gaussian distribution for each of the parameters. On every iteration of training, H-PPO runs by its policies $\pi_{\theta_d}$ and $\pi_{\theta_c}$ in the environment for $T$ timesteps and updates these two policies with the collected samples. The discrete policy $\pi_{\theta_d}$ and the continuous policy $\pi_{\theta_c}$ are updated separately by minimizing their respective clipped surrogate objective. The objective for the discrete policy $\pi_{\theta_d}$ is given by
\begin{equation}
\label{equ:hppo-objective-discrete}
\resizebox{.91\linewidth}{!}{$
    \displaystyle
    L^{\mathrm{CLIP}}_{d}(\theta_d) = \hat{\mathbb{E}}_t [\min(r_t^d(\theta_d)\hat{A}_t, \mathrm{clip}(r_t^d(\theta_d), 1-\epsilon,1+\epsilon)\hat{A}_t)],
$}
\end{equation}
and similarly the objective for the continuous policy $\pi_{\theta_c}$ is
\begin{equation}
\label{equ:hppo-objective-continuous}
\resizebox{.89\linewidth}{!}{$
    \displaystyle
    L^{\mathrm{CLIP}}_{c}(\theta_c) = \hat{\mathbb{E}}_t [\min(r_t^c(\theta_c)\hat{A}_t, \mathrm{clip}(r_t^c(\theta_c), 1-\epsilon,1+\epsilon)\hat{A}_t)].
$}
\end{equation}
Here the probability ratio $r_t^d(\theta_d)$ only considers the discrete policy and $r_t^c(\theta_c)$ only considers the continuous policy. That is to say, even though the two policies work with each other to decide the complete action, their objectives are not explicitly conditioned on each other. In other words, $\pi_{\theta_d}$ and $\pi_{\theta_c}$ are viewed as two separate distributions instead of a joint distribution in policy optimization. For example, if the complete action executed at timestep $t$ ($t \in [0, T]$) is denoted by $a_t = (a, x_{a})$, $r_t^d(\theta_d)$ is defined as $\frac{\pi_{\theta_d}(a \mid s_t)}{\pi_{\theta_d{\mathrm{(old)}}}(a \mid s_t)}$ and $r_t^c(\theta_c)$ is defined as $\frac{\pi_{\theta_c}(x_a \mid s_t)}{\pi_{\theta_c{\mathrm{(old)}}}(x_a \mid s_t)}$.

\subsection{Hybrid Actor-Critic Architecture for General Hierarchical Action Space} \label{subsec:general-hybrid-actor-critic}

\begin{figure}[t]
\centering
\includegraphics[width=0.8\linewidth]{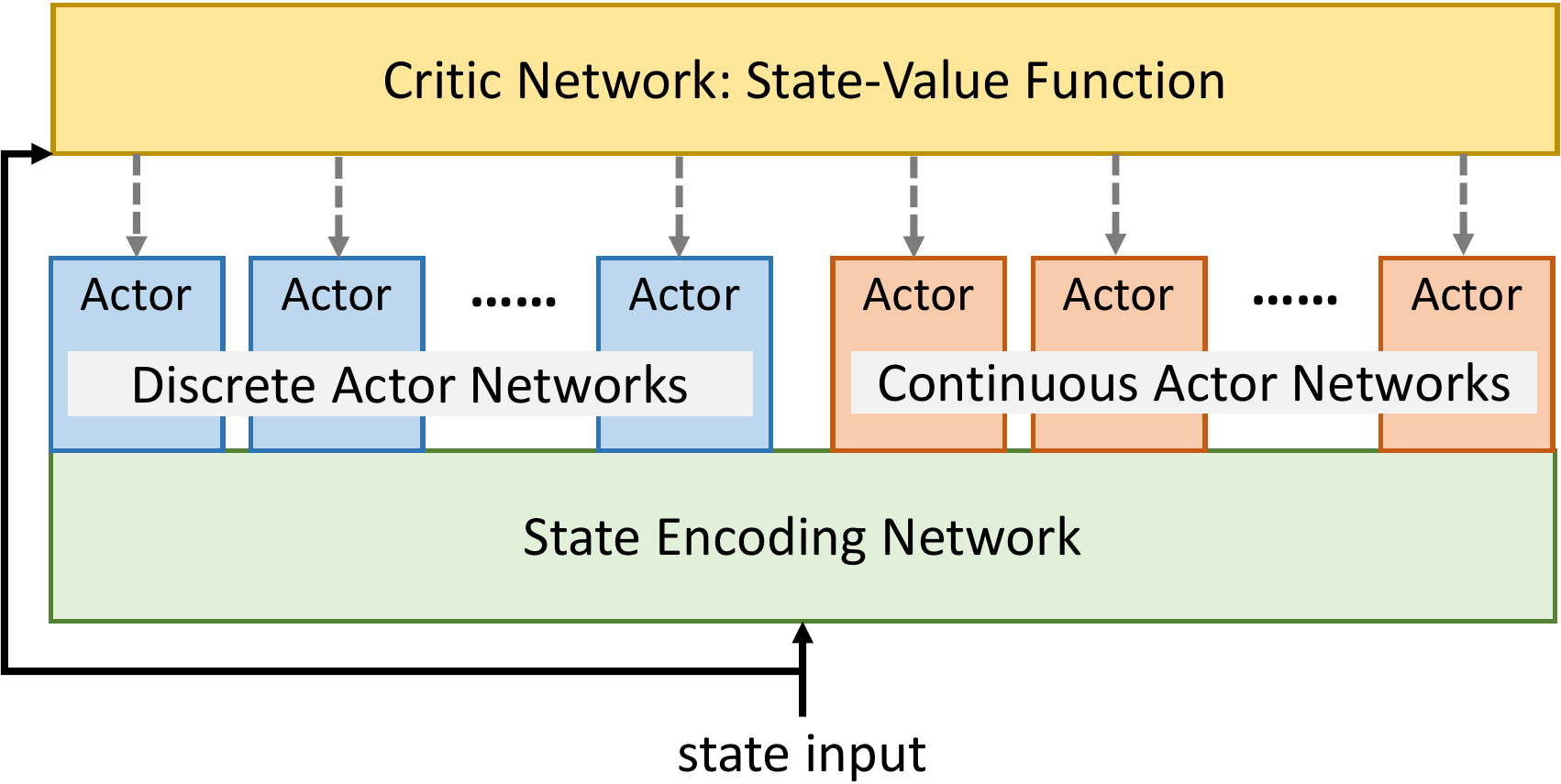}
\caption{Hybrid actor-critic architecture for general hierarchically structured action space.}
\label{fig:algorithm-structure-general}
\end{figure}

Apart from parameterized action space, the hybrid actor-critic architecture can be extended to solve RL problems with general hierarchical action space. Shown in Figure~\ref{fig:hierarchical-action-space}, the action-selection process in a general hierarchical action space could be represented with a tree structure. Each of the grey areas in Figure~\ref{fig:hierarchical-action-space} stands for an action-selection sub-problem. All internal nodes of the tree structure should be discrete action-selection sub-problems, and each discrete action on an internal node corresponds to an action-selection sub-problem of the next layer. The leaf nodes of the tree could be either discrete action-selection or continuous action-selection.

The hybrid actor-critic architecture for general hierarchical action space contains (see Figure~\ref{fig:algorithm-structure-general}) multiple parallel actor networks and one critic network. There is one actor network for each of the action-selection sub-problems, either discrete or continuous. The critic network here is the same as the critic in the hybrid actor-critic architecture for parameterized action space. The actor networks share the first few layers to encode the state and each of them generates either a stochastic discrete policy or a stochastic continuous policy. During training, the actors are updated as separate policies using a chosen policy optimization method such as PPO.

\section{Experiments} \label{sec:experiments}

\subsection{Environments}

\begin{figure}[t]
\centering
\subfigure[Catching Point]{
\fbox{\includegraphics[width=1.1in]{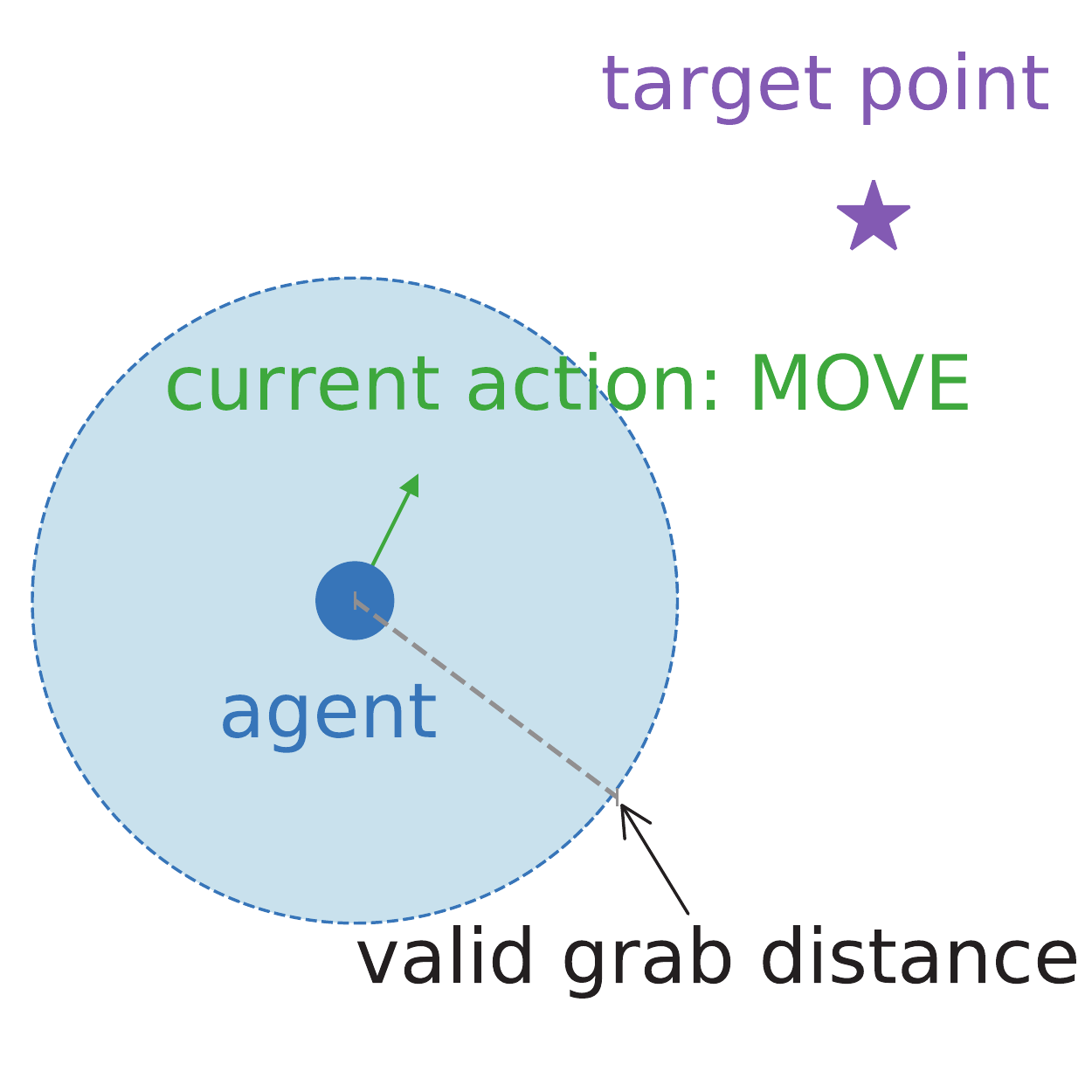}}
}
\subfigure[Moving]{
\fbox{\includegraphics[width=1.1in]{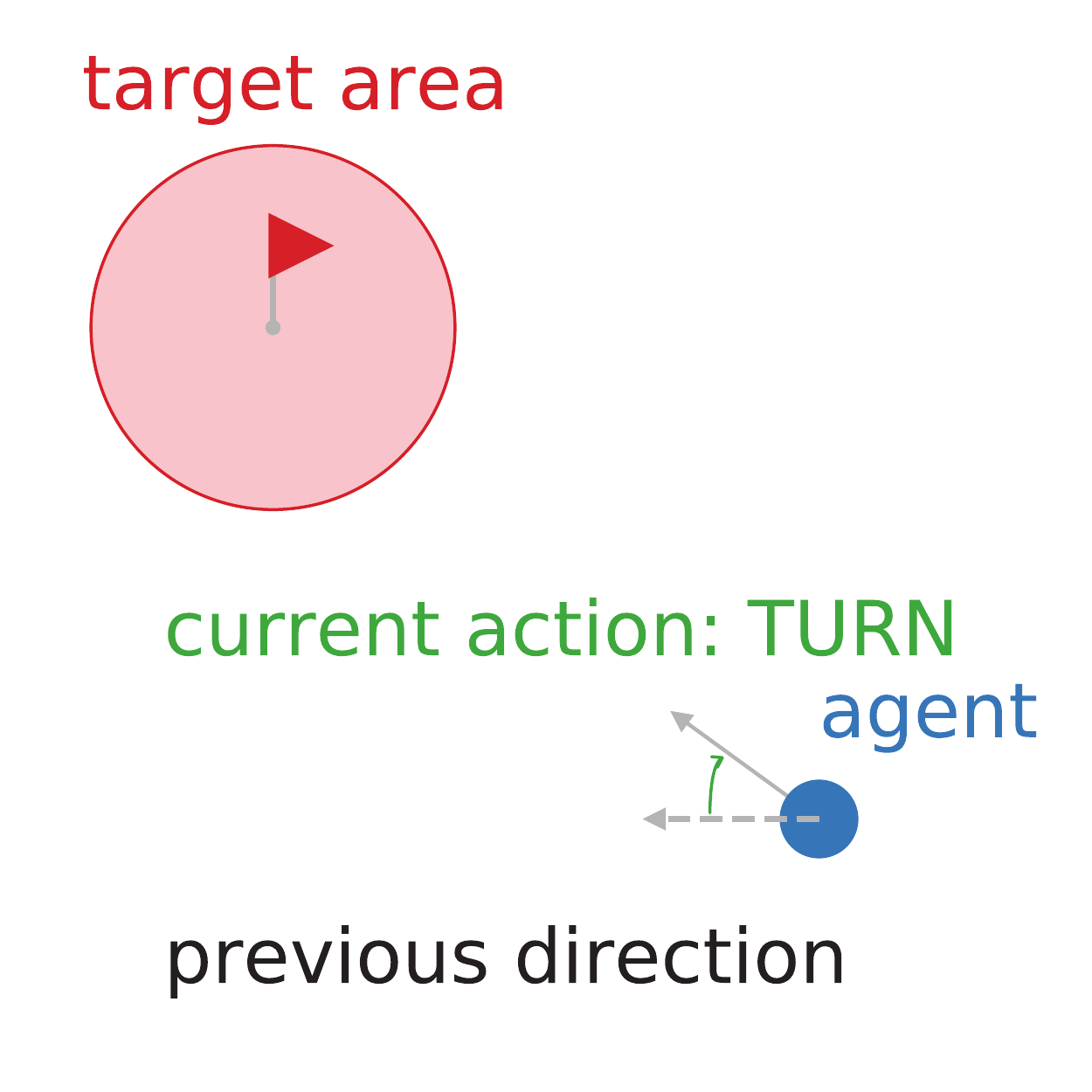}}
}
\subfigure[Chase and Attack]{
\fbox{\includegraphics[width=1.1in]{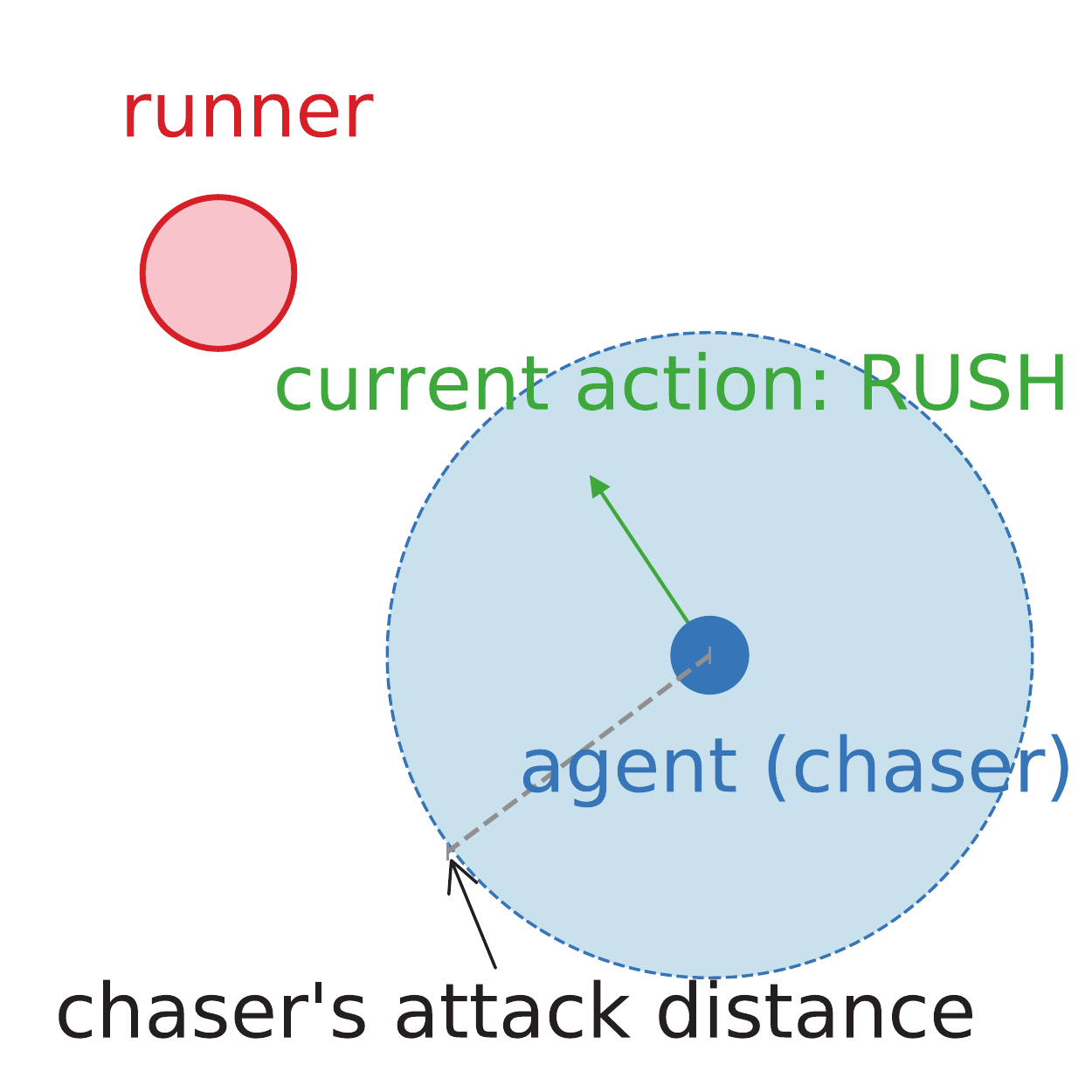}}
}
\subfigure[Half Field Football]{
\fbox{\includegraphics[width=1.1in]{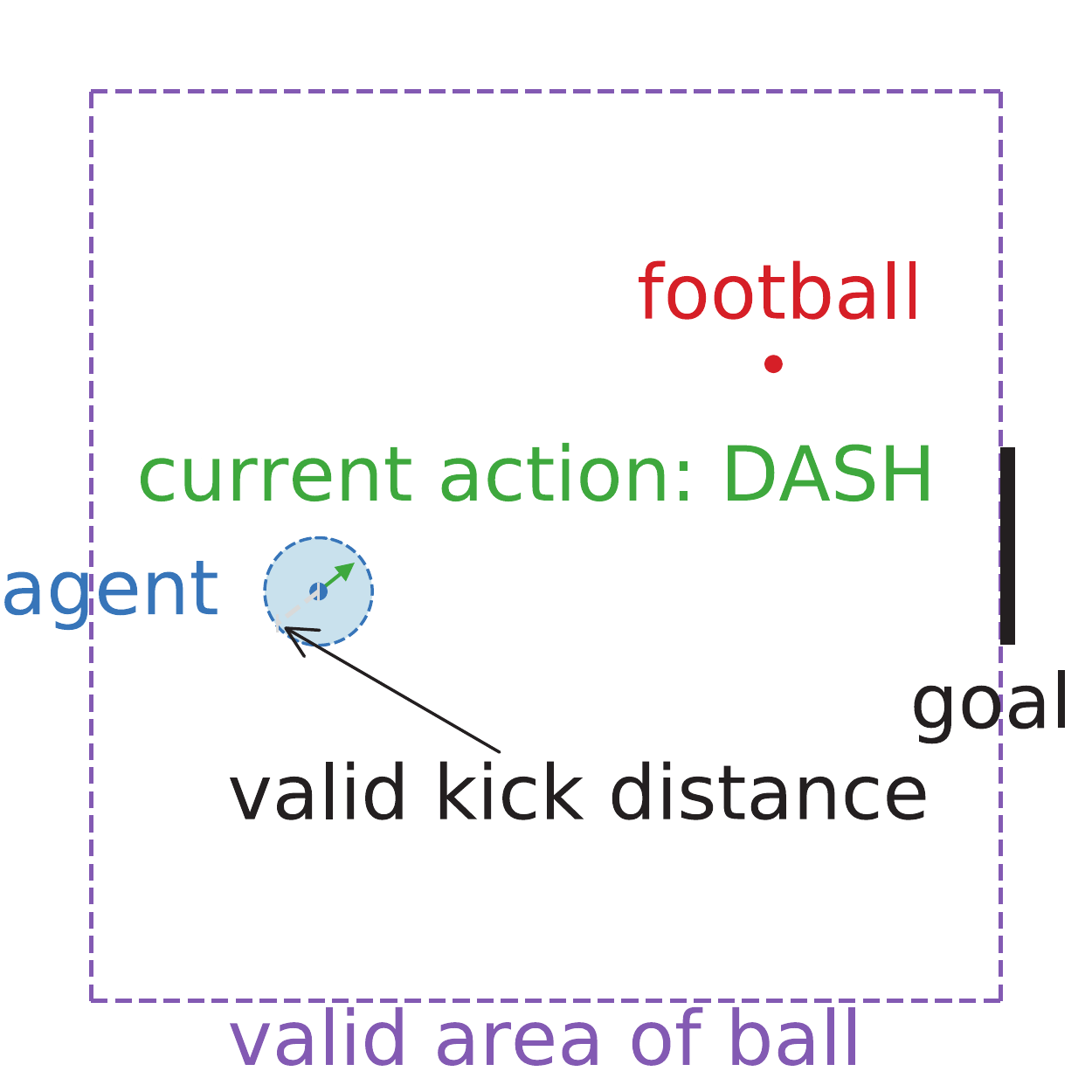}}
}
\caption{The four environments with parameterized action space used as the environments of the experiments. For clearness, the illustrations here show only a part of the whole field for the tasks except Football.}
\label{fig:environments}
\end{figure}

We create a collection of tasks with parameterized action space for the experiments, which is shown in Figure~\ref{fig:environments}. The tasks are briefly described below, and a more detailed description of the environment settings can be found in the supplemental material\footnote{\url{https://www.dropbox.com/s/s0ut449i3e2fsk1/suppl.pdf}}. Every task has a so-called "winning state", which is a final state of an episode indicating that the agent has succeeded in that episode.

\paragraph{Catching Point} In this task, the agent should catch a target point within limited chances. The parameterized actions are $\text{MOVE}(\text{direction}_\text{M})$ and CATCH, where MOVE means a movement of a constant distance in the specified direction and CATCH is an attempt to catch the target point. The agent could only try the CATCH action for up to $10$ times. The episode ends if the agent catches up the target point (the winning state), all of its $10$ chances ran out or the time limit exceeds.

\paragraph{Moving} In this scenario, the goal of the agent is to move towards a target area and stop in it. The agent can choose its action among $\text{ACCEL}(\text{power}_\text{A})$, $\text{TURN}(\text{direction}_\text{T})$ and $\text{BRAKE}$. The movement of the agent is always in the direction of its current direction. An episode ends if the agent stops in the target area (the winning state), it moves out of the field or the time limit exceeds.

\paragraph{Chase and Attack} In this task, the agent should chase a rule-based runner and attack it. The parameterized actions of the agent are $\text{RUSH}(\text{direction}_\text{R})$, $\text{ATTACK}(\text{direction}_\text{A})$. The runner has $3$ lives at the beginning, and it loses one life every time the agent performs a successful attack. The episode ends when the runner loses all of its lives (the winning state for the agent) or the time limit exceeds.

\paragraph{Half Field Football} This environment is a similar self-implemented version of a sub-scenario in the Half Field Offense (HFO)~\cite{hfo}. The task of the agent is to score a goal in the half football field with no goalie, and the same task in the original HFO environment is used as test environment for RL algorithms in parameterized action space by \citet{parameterized-ddpg} and \citet{pdqn}. The parameterized actions are $\text{DASH}(\text{power}_\text{D}, \text{direction}_\text{D})$, $\text{TURN}(\text{direction}_\text{T})$ and $\text{KICK}(\text{power}_\text{K}, \text{direction}_\text{K})$. The episode ends when the agent scores a goal (the winning state), the ball is out of the valid area, or the time limit exceeds.

\subsection{Experiment Settings and Results}
We evaluated the performance of the H-PPO on the four tasks above. In addition, we also implemented and tested the following three baseline algorithms: the extended DDPG for parameterized action space by \citet{parameterized-ddpg}, the P-DQN algorithm~\cite{pdqn} and DQN which first discretizes the parameterized action space.

The networks in the four algorithms are of the same size, and the hidden layer sizes for each network is $(256, 256, 128, 64)$. The replay buffer size for DDPG and DQN is $10000$, and the batch size for sampling is $32$. For DQN, we discretizes the action space of Chase and Attack into 30 actions, 16 discrete actions for Catching Point and 23 discrete actions for Moving. However, since the action space of Half Field Football task contains more parameters, the discretized action space has a relatively large size of $104$ even if we only discretize each direction parameter into $8$ values and each power parameter into $6$ values. 

\begin{figure}[t!]
\centering
\subfigure[Catching Point]{
\begin{minipage}[b]{0.5\textwidth}
\includegraphics[width=1.65in]{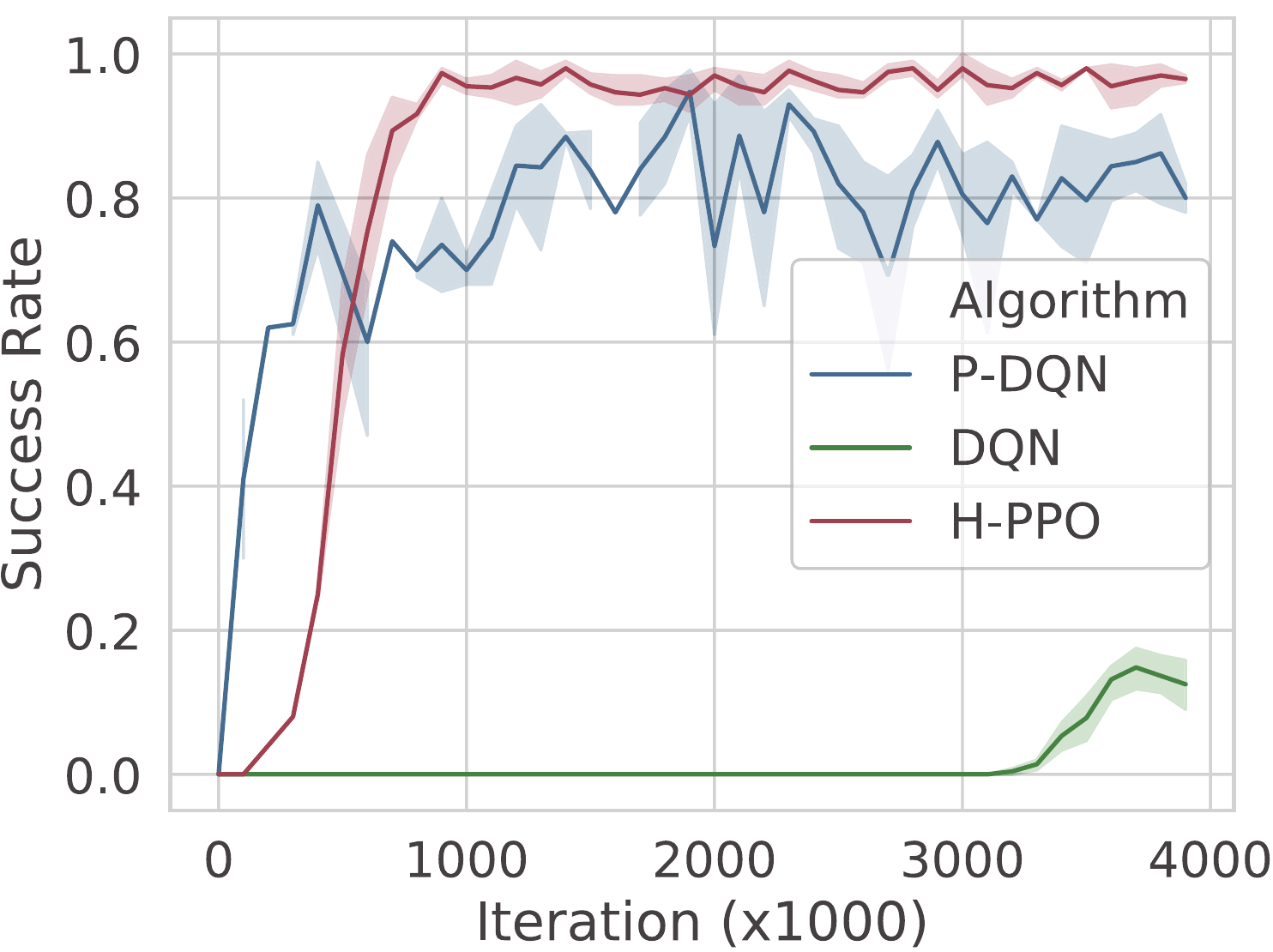}
\includegraphics[width=1.65in]{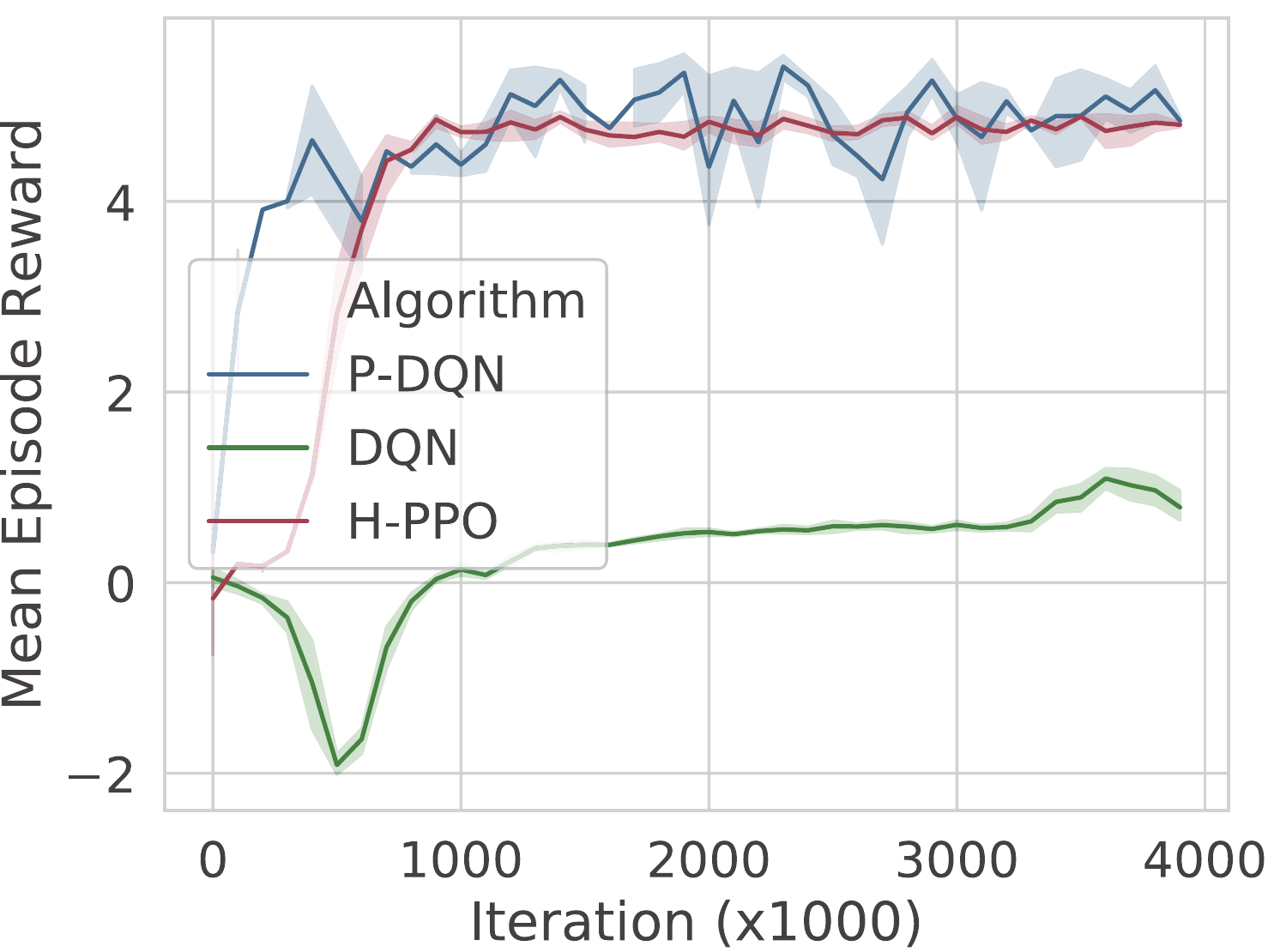}
\end{minipage}
}
\subfigure[Moving]{
\begin{minipage}[b]{0.5\textwidth}
\includegraphics[width=1.65in]{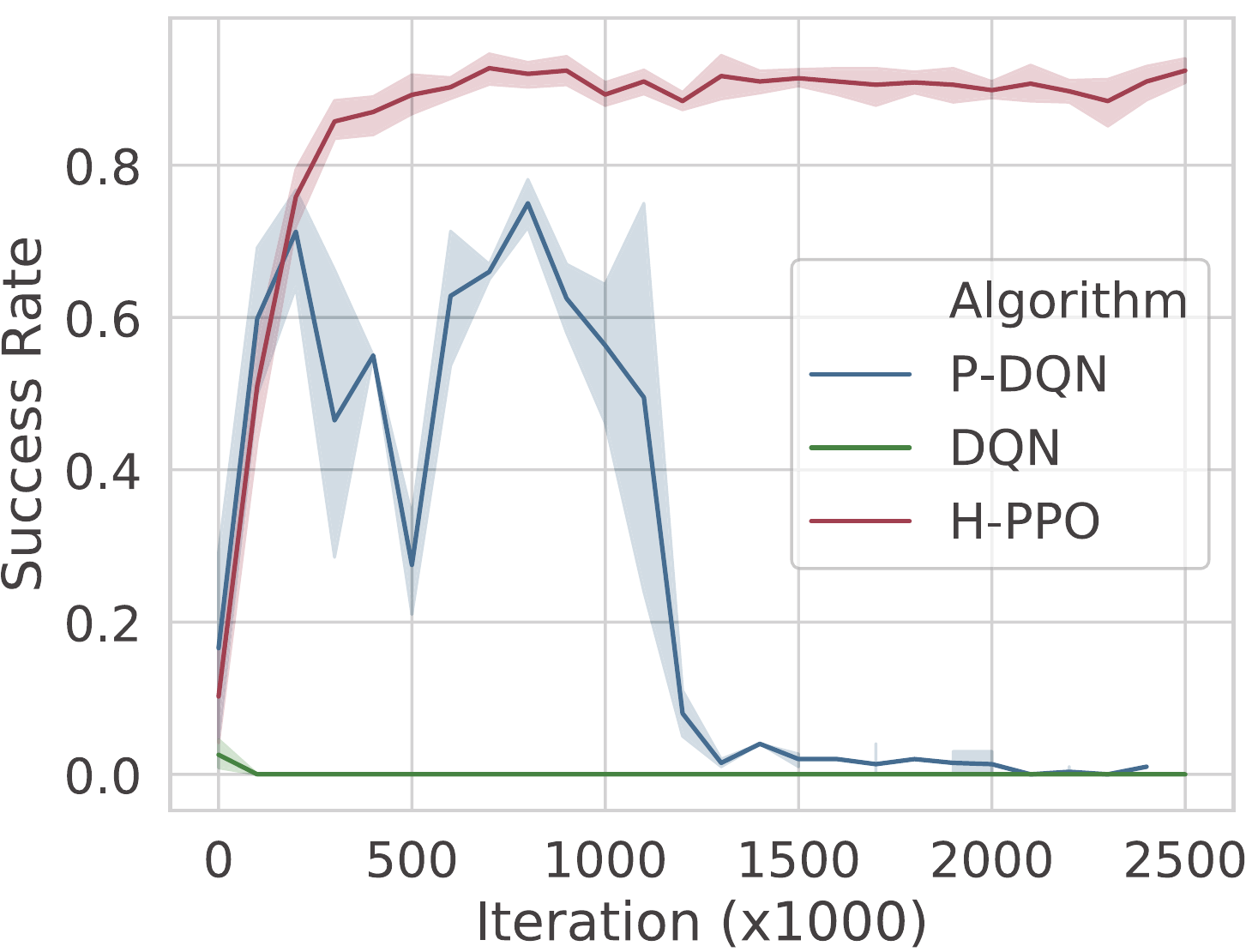}
\includegraphics[width=1.65in]{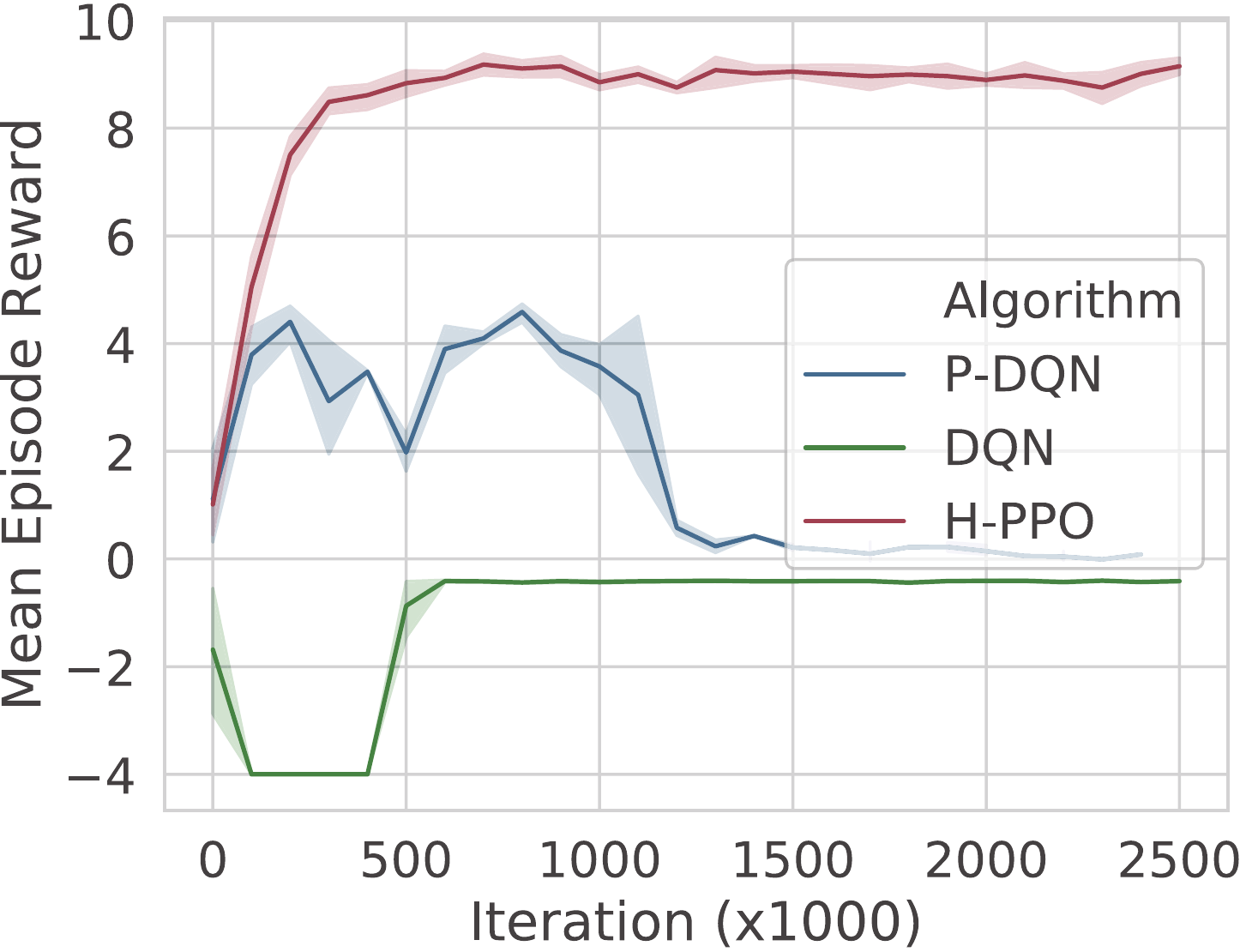}
\end{minipage}
}
\subfigure[Chase and Attack]{
\begin{minipage}[b]{0.5\textwidth}
\includegraphics[width=1.65In]{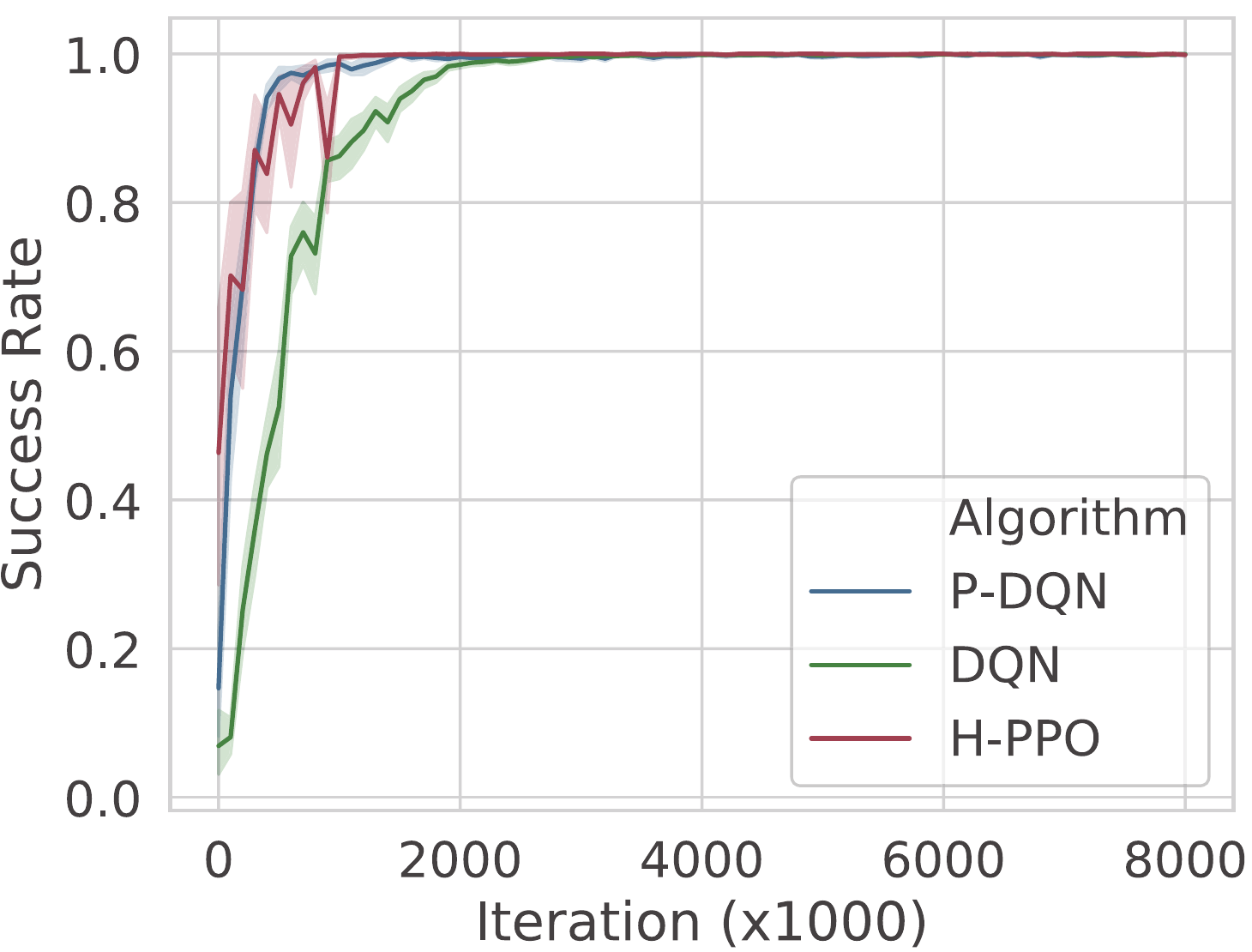}
\includegraphics[width=1.65in]{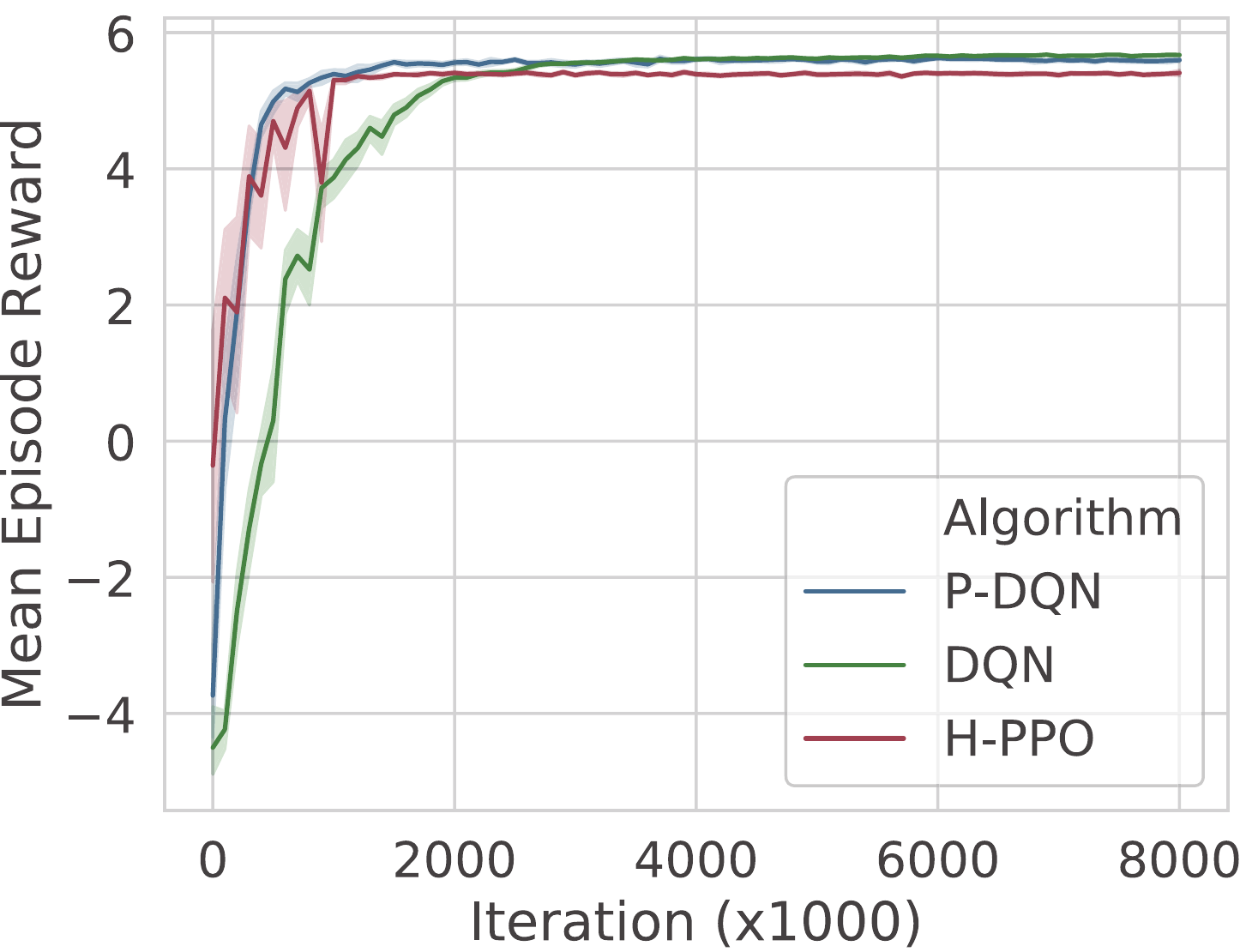}
\end{minipage}
}
\subfigure[Half Field Football]{
\begin{minipage}[b]{0.5\textwidth}
\includegraphics[width=1.65in]{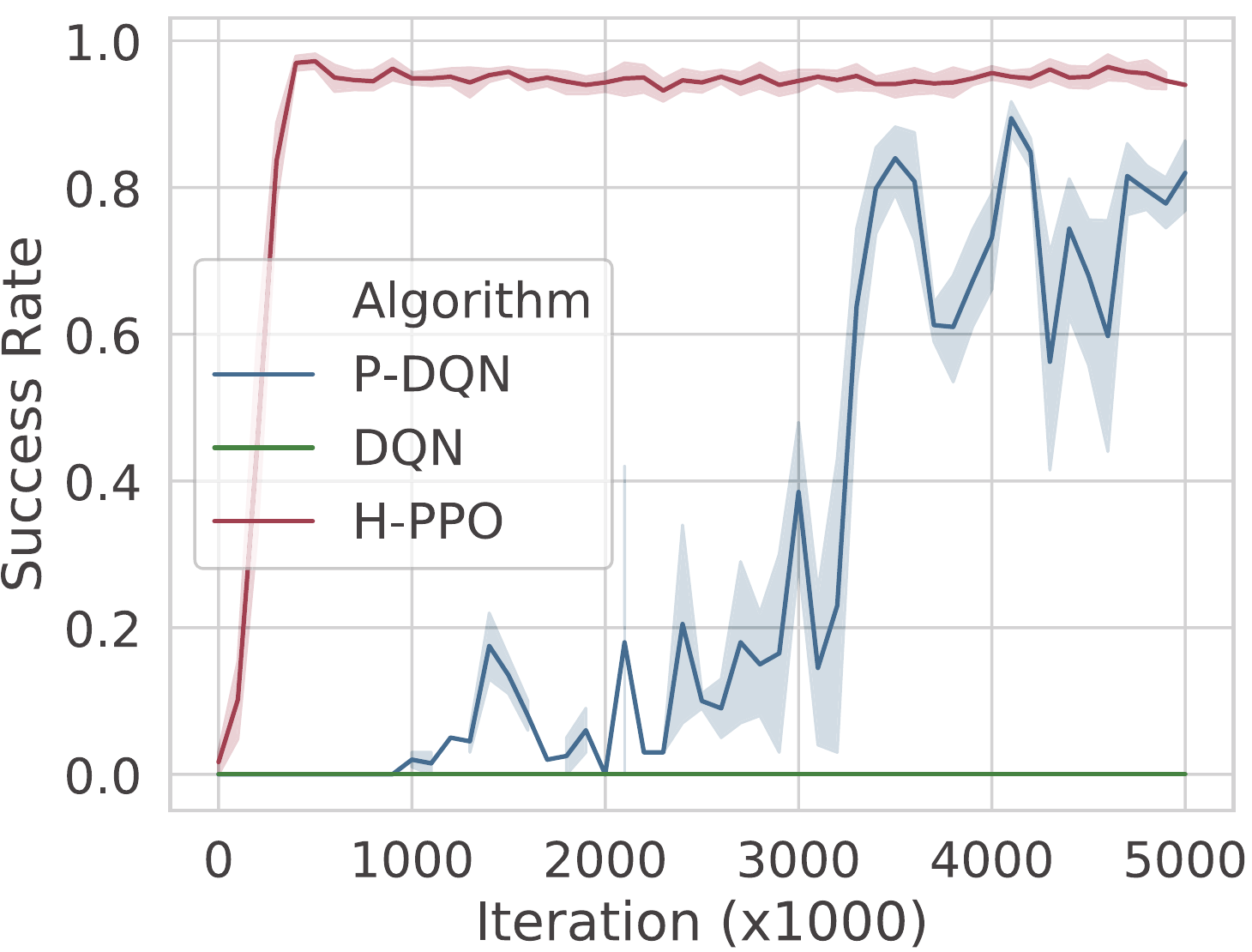}
\includegraphics[width=1.65in]{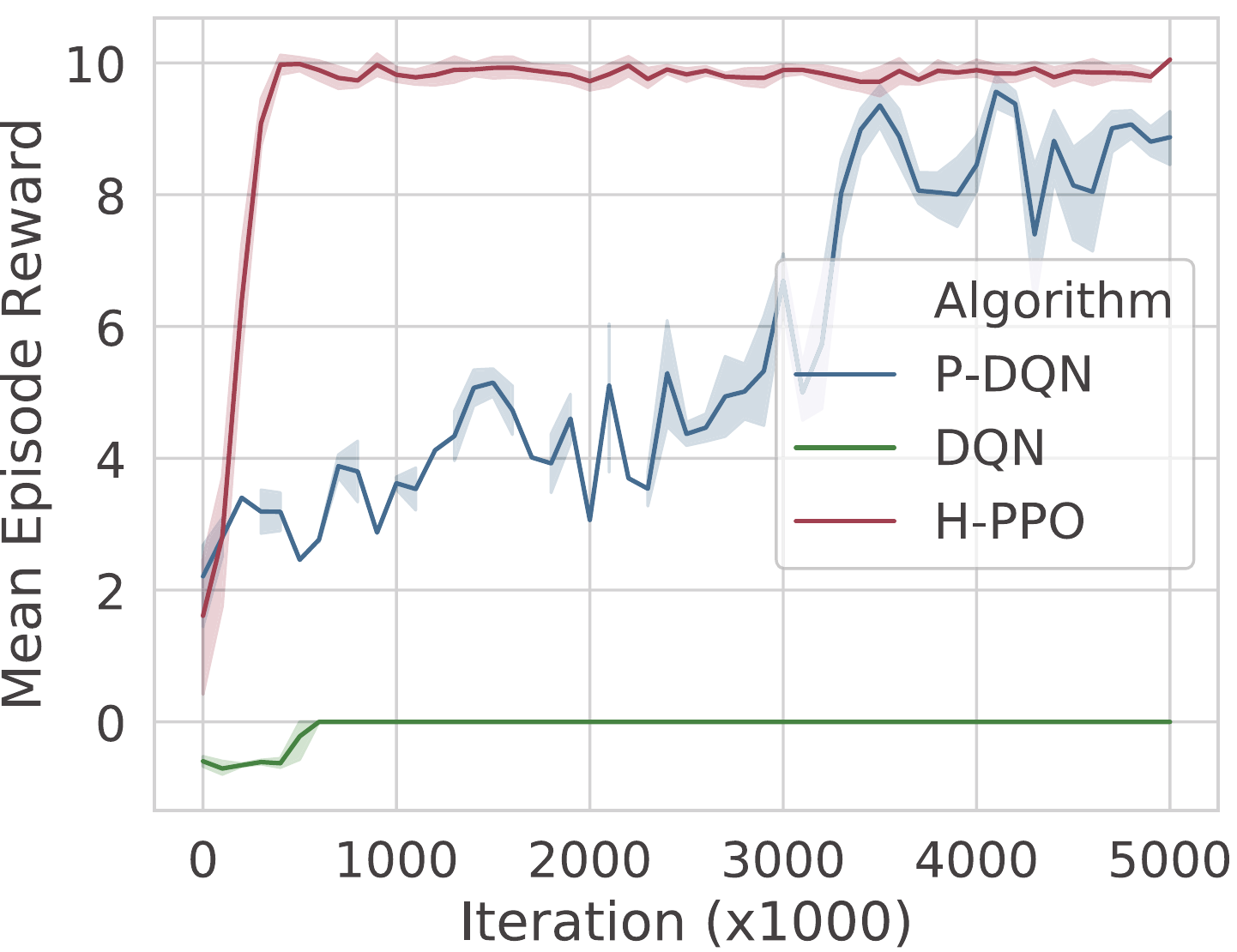}
\end{minipage}
}
\caption{Results of the experiments on the four tasks. Left column: Success rate of each method during training. Right column: Mean episode reward of each method during training.}
\label{fig:results}
\end{figure}

\begin{table}[b]
\resizebox{\columnwidth}{!}{%
\begin{tabular}{@{}llrrr@{}}
\toprule
Environment                          & Algorithm & SuccRate & SD of SuccRate & Mean Reward \\ \midrule
\multirow{3}{*}{Catching Point}      & DQN       & $6.13\%$        & $\pm8.21\%$             & 0.796           \\
                                     & P-DQN       & $82.52\%$        & $\pm11.60\%$             & 4.977           \\
                                     & \bf{H-PPO}       & $\mathbf{96.32\%}$        & $\mathbf{\pm4.82\%}$             & $\mathbf{4.790}$           \\ \midrule
\multirow{3}{*}{Moving}              & DQN       & $0.00\%$        & $\pm0.00\%$             & -0.415           \\
                                     & P-DQN       & $1.56\%$        & $\pm2.78\%$             & 0.173          \\
                                     & \bf{H-PPO}       & $\mathbf{90.45\%}$        & $\mathbf{\pm6.75\%}$             & $\mathbf{8.955}$       \\ \midrule
\multirow{3}{*}{Chase and Attack}    & DQN       & $99.91\%$        & $\pm0.74\%$             & 5.664          \\
                                     & P-DQN       & $99.85\%$        & $\pm0.84\%$             & 5.589          \\
                                     & \bf{H-PPO}       & $\mathbf{99.98\%}$        & $\mathbf{\pm0.30\%}$             & $\mathbf{5.393}$        \\ \midrule
\multirow{3}{*}{Half Field Football} & DQN       & $0.00\%$        & $\pm0.00\%$             & 0.000         \\
                                     & P-DQN       & $76.31\%$        & $\pm16.81\%$             & 8.762           \\
                                     & \bf{H-PPO}       & $\mathbf{95.39\%}$        & $\mathbf{\pm4.81\%}$             & $\mathbf{9.849}$        \\ \bottomrule
\end{tabular}%
}
\caption{Success rate, standard deviation of success rate and mean episode reward achieved by DQN, P-DQN and H-PPO in the experiment environments.}
\label{tab:results}
\end{table}

Figure~\ref{fig:results} shows the experiments results, which contains both the success rate (the percentage of episodes which ends in the winning state) and mean episode reward during training of the methods in the four test environments. The experiment results of DDPG are not included here because the performance of DDPG in our experiments was far worse than in the original paper, demonstrated a large variance and it failed to learn reasonable policies, which is an issue also reported by \citet{hierarchical-approach-for-parl}. Table~\ref{tab:results} shows the success rate, standard deviation of success rate and mean episode reward achieved by DQN, P-DQN and H-PPO after the same number of iterations of learning in the four environments. As we can see from the results, H-PPO showed stable learning and achieved high success rate on all the four tasks. Moreover, H-PPO outperformed other methods by a large margin in three of the four environments (except in Chase and Attack, where the three algorithms all achieved similar success rate and H-PPO had the lowest variance, see Table~\ref{tab:results}). It generally achieved higher success rate, faster convergence and lower variance than other methods in the experiments.

\begin{figure}[t]
\centering
\subfigure[TURN]{
\fbox{\includegraphics[width=0.7in]{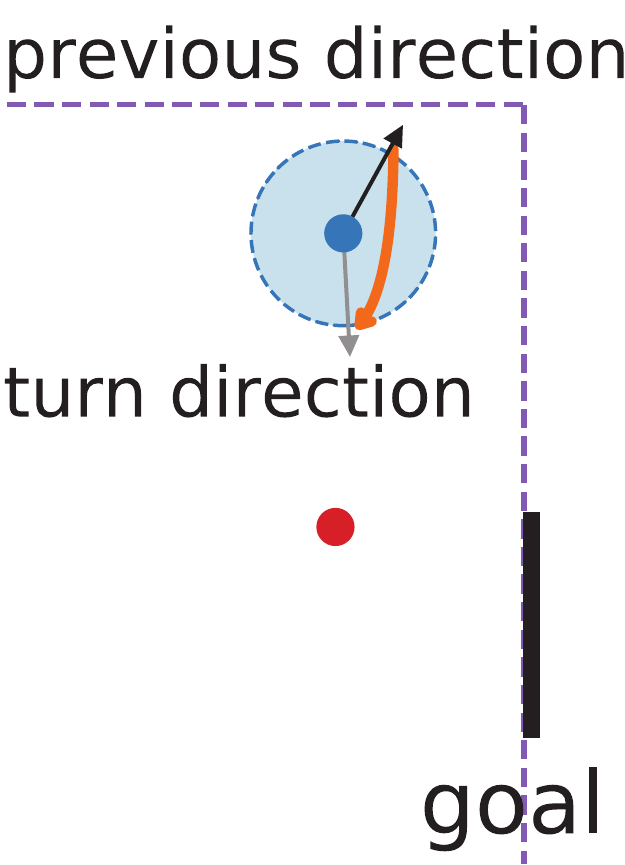}}
}
\subfigure[DASH]{
\fbox{\includegraphics[width=0.7in]{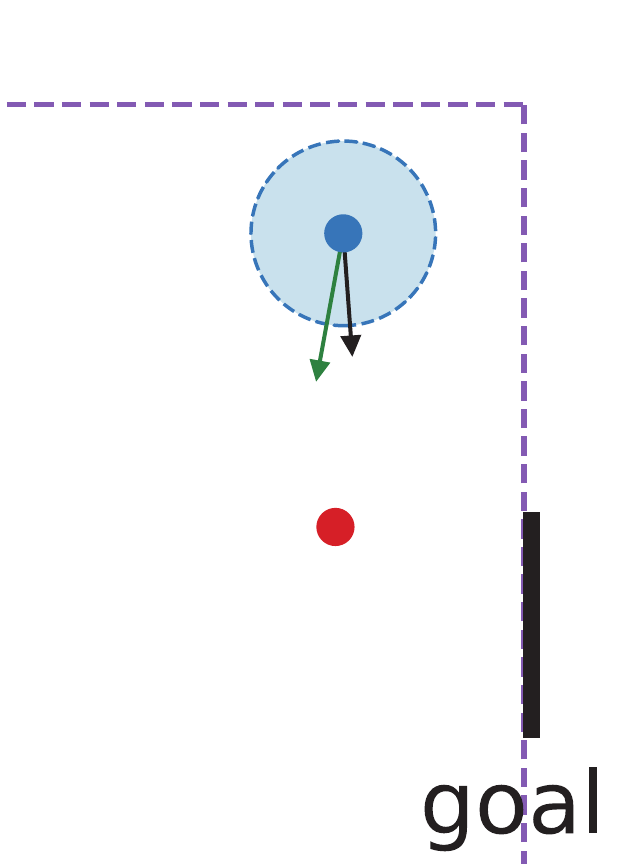}}
}
\subfigure[KICK]{
\fbox{\includegraphics[width=0.7in]{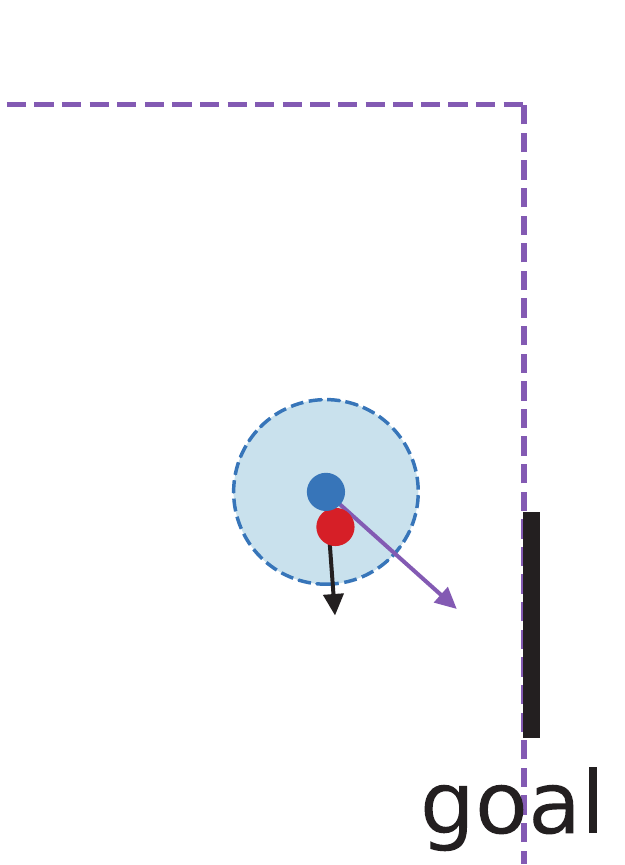}}
}
\caption{Illustration of the parameterized actions selected by the H-PPO agent in three frames of a Half Field Football episode.}
\label{fig:actions}
\end{figure}

To illustrate the learned action-selection and parameter-selection policy of H-PPO from the micro perspective, Figure~\ref{fig:actions} shows the states of three frames (placed in the order of time) observed in an episode of Half Field Football and the parameterized actions selected by the H-PPO agent in these frames. Black arrows in Figure~\ref{fig:actions} indicates the facing direction of the agent, and arrows in other colors show the selected actions and parameters. The agent was not close to the ball in frame (a) and its direction was not toward the ball, so it performed a TURN (the discrete action) with the specified angle (the continuous parameter, shown in orange) to get closer to the ball. Then it was roughly facing the ball in frame (b), and it chose DASH along with proper power and direction (shown in green). Finally, it took a KICK in the direction (shown in purple) toward the goal in frame (c), when it was close enough to the ball to perform the KICK. This example shows the capability of H-PPO to learn the discrete action-selection policy and the continuous parameter-selection policy in coordination with its hybrid actor-critic architecture. This coordination during training comes from the characteristics of H-PPO that not only the two parallel actors share the first few layers of their networks, they also share the same critic to perform policy optimization updates with similar objectives given by Eq.~(\ref{equ:hppo-objective-discrete}) and Eq.~(\ref{equ:hppo-objective-continuous}).

\section{Conclusion and Future Work} \label{sec:conclusion}

This paper introduced a hybrid actor-critic architecture for reinforcement learning in parameterized action space where the discrete action policy and the continuous parameter policy are trained in parallel as separate actors with a global critic. As the hybrid actor-critic architecture is flexible in the choice of the policy optimization method, we also presented H-PPO, which is an implementation of the architecture based on PPO. Empirically, H-PPO achieves stable learning in all of the four tasks with parameterized action space and outperforms previous methods of parameterized action reinforcement learning.

Although this paper is mainly focused on reinforcement learning in parameterized action space, we also briefly presented an extended version of the hybrid actor-critic architecture for general hierarchical action spaces. More experiments are needed to test the performance of this architecture in general hierarchical action spaces, and we leave this investigation as future work.

\section*{Acknowledgments}

This work is supported by Tencent AI Lab Joint Research Program. The corresponding author Weinan Zhang thanks the support of National Natural Science Foundation of China (61702327, 61772333, 61632017) and Shanghai Sailing Program (17YF1428200).

\bibliographystyle{named}
\bibliography{ijcai19}

\begin{thebibliography}{}

\bibitem[\protect\citeauthoryear{Hasselt \bgroup \em et al.\egroup
  }{2016}]{double-dqn}
Hado~van Hasselt, Arthur Guez, and David Silver.
\newblock Deep reinforcement learning with double q-learning.
\newblock In {\em Proceedings of the Thirtieth AAAI Conference on Artificial
  Intelligence}, AAAI'16. AAAI Press, 2016.

\bibitem[\protect\citeauthoryear{Hausknecht and
  Stone}{2016}]{parameterized-ddpg}
Matthew Hausknecht and Peter Stone.
\newblock Deep reinforcement learning in parameterized action space.
\newblock In {\em Proceedings of the International Conference on Learning
  Representations (ICLR)}, May 2016.

\bibitem[\protect\citeauthoryear{Hausknecht \bgroup \em et al.\egroup
  }{2016}]{hfo}
Matthew Hausknecht, Prannoy Mupparaju, Sandeep Subramanian, Shivaram
  Kalyanakrishnan, and Peter Stone.
\newblock Half field offense: An environment for multiagent learning and ad hoc
  teamwork.
\newblock In {\em AAMAS Adaptive Learning Agents (ALA) Workshop}, May 2016.

\bibitem[\protect\citeauthoryear{Heess \bgroup \em et al.\egroup }{2015}]{svg}
Nicolas Heess, Gregory Wayne, David Silver, Timothy Lillicrap, Tom Erez, and
  Yuval Tassa.
\newblock Learning continuous control policies by stochastic value gradients.
\newblock In C.~Cortes, N.~D. Lawrence, D.~D. Lee, M.~Sugiyama, and R.~Garnett,
  editors, {\em Advances in NIPS 28}. Curran Associates, Inc., 2015.

\bibitem[\protect\citeauthoryear{Kober \bgroup \em et al.\egroup
  }{2013}]{rl-robotics-survey}
Jens Kober, J.~Andrew Bagnell, and Jan Peters.
\newblock Reinforcement learning in robotics: A survey.
\newblock {\em The International Journal of Robotics Research},
  32(11):1238--1274, 2013.

\bibitem[\protect\citeauthoryear{Konda and Tsitsiklis}{2000}]{actor-critic}
Vijay~R. Konda and John~N. Tsitsiklis.
\newblock Actor-critic algorithms.
\newblock In S.~A. Solla, T.~K. Leen, and K.~M\"{u}ller, editors, {\em Advances
  in Neural Information Processing Systems 12}, pages 1008--1014. MIT Press,
  2000.

\bibitem[\protect\citeauthoryear{Lillicrap \bgroup \em et al.\egroup
  }{2016}]{ddpg}
Timothy~P. Lillicrap, Jonathan~J. Hunt, Alexander Pritzel, Nicolas Heess, Tom
  Erez, Yuval Tassa, David Silver, and Daan Wierstra.
\newblock Continuous control with deep reinforcement learning.
\newblock {\em CoRR}, abs/1509.02971, 2016.

\bibitem[\protect\citeauthoryear{Masson \bgroup \em et al.\egroup
  }{2016}]{q-pamdp}
Warwick Masson, Pravesh Ranchod, and George Konidaris.
\newblock Reinforcement learning with parameterized actions.
\newblock In {\em Proceedings of the Thirtieth AAAI Conference on Artificial
  Intelligence}, AAAI'16, pages 1934--1940. AAAI Press, 2016.

\bibitem[\protect\citeauthoryear{Mnih \bgroup \em et al.\egroup
  }{2013}]{mnih-atari-2013}
Volodymyr Mnih, Koray Kavukcuoglu, David Silver, Alex Graves, Ioannis
  Antonoglou, Daan Wierstra, and Martin Riedmiller.
\newblock Playing atari with deep reinforcement learning.
\newblock In {\em NIPS Deep Learning Workshop}. 2013.

\bibitem[\protect\citeauthoryear{Mnih \bgroup \em et al.\egroup
  }{2016}]{async-drl}
Volodymyr Mnih, Adria~Puigdomenech Badia, Mehdi Mirza, Alex Graves, Timothy
  Lillicrap, Tim Harley, David Silver, and Koray Kavukcuoglu.
\newblock Asynchronous methods for deep reinforcement learning.
\newblock In {\em Proceedings of The 33rd International Conference on Machine
  Learning}, Proceedings of Machine Learning Research, New York, New York, USA,
  2016. PMLR.

\bibitem[\protect\citeauthoryear{OpenAI}{2018}]{openai5}
OpenAI.
\newblock Openai five.
\newblock https://blog.openai.com/openai-five/, 6 2018.

\bibitem[\protect\citeauthoryear{Schulman \bgroup \em et al.\egroup
  }{2015}]{trpo}
John Schulman, Sergey Levine, Pieter Abbeel, Michael Jordan, and Philipp
  Moritz.
\newblock Trust region policy optimization.
\newblock In Francis Bach and David Blei, editors, {\em Proceedings of the 32nd
  International Conference on Machine Learning}, Proceedings of Machine
  Learning Research, Lille, France, 2015. PMLR.

\bibitem[\protect\citeauthoryear{Schulman \bgroup \em et al.\egroup
  }{2017}]{ppo}
John Schulman, Filip Wolski, Prafulla Dhariwal, Alec Radford, and Oleg Klimov.
\newblock Proximal policy optimization algorithms.
\newblock {\em CoRR}, abs/1707.06347, 2017.

\bibitem[\protect\citeauthoryear{Sherstov and Stone}{2005}]{tile-coding}
Alexander~A. Sherstov and Peter Stone.
\newblock Function approximation via tile coding: Automating parameter choice.
\newblock In Jean-Daniel Zucker and Lorenza Saitta, editors, {\em Abstraction,
  Reformulation and Approximation}, pages 194--205, Berlin, Heidelberg, 2005.
  Springer Berlin Heidelberg.

\bibitem[\protect\citeauthoryear{Silver \bgroup \em et al.\egroup }{2014}]{dpg}
David Silver, Guy Lever, Nicolas Heess, Thomas Degris, Daan Wierstra, and
  Martin Riedmiller.
\newblock Deterministic policy gradient algorithms.
\newblock In {\em Proceedings of the 31st International Conference on
  International Conference on Machine Learning - Volume 32}, ICML'14, pages
  I--387--I--395. JMLR.org, 2014.

\bibitem[\protect\citeauthoryear{Silver \bgroup \em et al.\egroup
  }{2016}]{alphago}
David Silver, Aja Huang, Christopher~J. Maddison, et~al.
\newblock Mastering the game of go with deep neural networks and tree search.
\newblock {\em Nature}, 529:484--503, 2016.

\bibitem[\protect\citeauthoryear{Silver \bgroup \em et al.\egroup
  }{2017}]{alphago-2}
David Silver, Julian Schrittwieser, Karen Simonyan, et~al.
\newblock Mastering the game of go without human knowledge.
\newblock {\em Nature}, 550:354--, October 2017.

\bibitem[\protect\citeauthoryear{Sutton \bgroup \em et al.\egroup
  }{2000}]{policy-gradient}
Richard~S Sutton, David~A. McAllester, Satinder~P. Singh, and Yishay Mansour.
\newblock Policy gradient methods for reinforcement learning with function
  approximation.
\newblock In S.~A. Solla, T.~K. Leen, and K.~M\"{u}ller, editors, {\em Advances
  in Neural Information Processing Systems 12}, pages 1057--1063. MIT Press,
  2000.

\bibitem[\protect\citeauthoryear{Vinyals \bgroup \em et al.\egroup
  }{2017}]{star-craft-2}
Oriol Vinyals, Timo Ewalds, Sergey Bartunov, et~al.
\newblock Starcraft {II:} {A} new challenge for reinforcement learning.
\newblock {\em CoRR}, abs/1708.04782, 2017.

\bibitem[\protect\citeauthoryear{Wang \bgroup \em et al.\egroup
  }{2016}]{dueling-q}
Ziyu Wang, Tom Schaul, Matteo Hessel, Hado Hasselt, Marc Lanctot, and Nando
  Freitas.
\newblock Dueling network architectures for deep reinforcement learning.
\newblock In {\em Proceedings of The 33rd International Conference on Machine
  Learning}, Proceedings of Machine Learning Research, New York, New York, USA,
  2016. PMLR.

\bibitem[\protect\citeauthoryear{Watkins and Dayan}{1992}]{q-learning}
Christopher J. C.~H. Watkins and Peter Dayan.
\newblock Q-learning.
\newblock In {\em Machine Learning}, pages 279--292, 1992.

\bibitem[\protect\citeauthoryear{Wei \bgroup \em et al.\egroup
  }{2018}]{hierarchical-approach-for-parl}
Ermo Wei, Drew Wicke, and Sean Luke.
\newblock Hierarchical approaches for reinforcement learning in parameterized
  action space.
\newblock {\em CoRR}, abs/1810.09656, 2018.

\bibitem[\protect\citeauthoryear{Xiong \bgroup \em et al.\egroup }{2018}]{pdqn}
Jiechao Xiong, Qing Wang, Zhuoran Yang, Peng Sun, Lei Han, Yang Zheng, Haobo
  Fu, Tong Zhang, Ji~Liu, and Han Liu.
\newblock Parametrized deep q-networks learning: Reinforcement learning with
  discrete-continuous hybrid action space.
\newblock {\em CoRR}, abs/1810.06394, 2018.

\end{thebibliography}

\end{document}